\documentclass[letterpaper]{article} 
\usepackage[]{aaai2026}  
\usepackage{times}  
\usepackage{helvet}  
\usepackage{courier}  
\usepackage[hyphens]{url}  
\usepackage{graphicx} 
\usepackage{natbib}  
\usepackage{caption} 
\usepackage{algorithm}
\usepackage{algorithmic}
\usepackage{amsmath}
\usepackage{amssymb}
\usepackage{csquotes}
\MakeOuterQuote{"}
\usepackage{siunitx}
\usepackage{multirow}
\usepackage{booktabs}
\usepackage{array}
\usepackage{bm} 
\usepackage{makecell}

\usepackage{newfloat}
\usepackage{listings}
\DeclareCaptionStyle{ruled}{labelfont=normalfont,labelsep=colon,strut=off} 
\floatstyle{ruled}
\newfloat{listing}{tb}{lst}{}
\floatname{listing}{Listing}
\title{MechaFormer: Sequence Learning for Kinematic Mechanism Design Automation}

\author {
    Diana Bolanos \textsuperscript{\rm 1},
    Mohammadmehdi Ataei\textsuperscript{\rm 1},
    Pradeep Kumar Jayaraman\textsuperscript{\rm 1}
}
\affiliations {
    \textsuperscript{\rm 1}Autodesk Research\\
    dbolanos@berkeley.edu, mehdi.ataei@autodesk.com, pradeep.kumar.jayaraman@autodesk.com
}

\begin{document}

\maketitle

\begin{abstract}
Designing mechanical mechanisms to trace specific paths is a classic yet notoriously difficult engineering problem, characterized by a vast and complex search space of discrete topologies and continuous parameters. We introduce MechaFormer, a Transformer-based model that tackles this challenge by treating mechanism design as a conditional sequence generation task. Our model learns to translate a target curve into a domain-specific language (DSL) string, simultaneously determining the mechanism's topology and geometric parameters in a single, unified process. MechaFormer significantly outperforms existing baselines, achieving state-of-the-art path-matching accuracy and generating a wide diversity of novel and valid designs. We demonstrate a suite of sampling strategies that can dramatically improve solution quality and offer designers valuable flexibility. Furthermore, we show that the high-quality outputs from MechaFormer serve as excellent starting points for traditional optimizers, creating a hybrid approach that finds superior solutions with remarkable efficiency.

\end{abstract}

\begin{figure*}[t]
    \centering
    \includegraphics[width=0.9\linewidth]{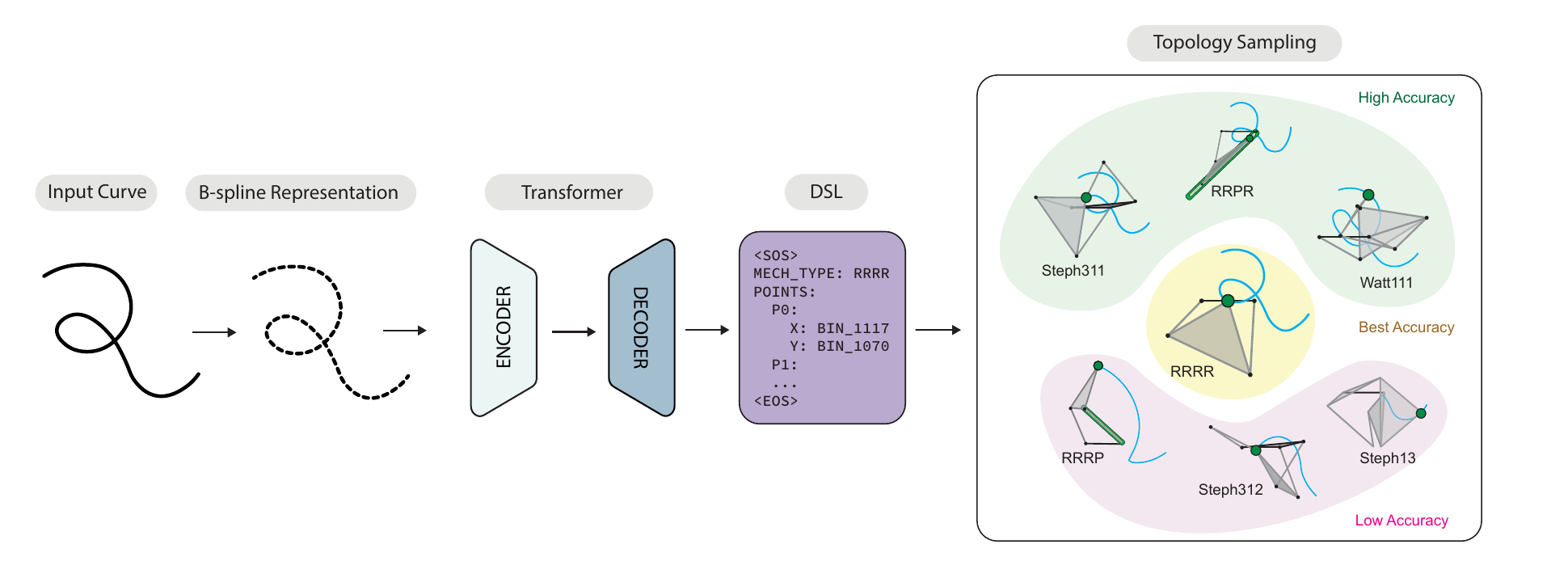}
    \caption{Mechanism-sampling pipeline. The input curve is converted to a B-spline, encoded by a Transformer, and decoded into a DSL that specifies mechanism topology and joint parameters. Topology sampling then generates and evaluates candidates, with clusters indicating accuracy.}
    \label{fig:fig1}
\end{figure*}

\section{Introduction}
Mechanism synthesis is a foundational problem in mechanical design that involves the systematic generation and selection of feasible kinematic topologies and configurations to achieve specific motion or functional requirements \cite{hartenberg1964kinematic, angeles2003fundamentals}. It plays a crucial role in various applications, from industrial robotics and manufacturing machinery to biomedical devices and consumer electronics. The overarching goal of mechanism synthesis is to determine the best arrangement of joints and links such that a mechanical system performs a desired task reliably and efficiently, often under constraints like space limitations, joint limits, or load-bearing capacities \cite{norton2007design}. A designer must select a topology (the number and arrangement of links and joints) from a discrete set of graphs while simultaneously optimizing continuous geometric parameters like joint locations. This mixed discrete-continuous space, combined with a highly non-linear and non-convex relationship between parameters and motion, is commonly encountered in mechanical design problems \cite{bolanos2023selecting,martins2021engineering}, and makes it notoriously hard for analytical and optimization-based methods.

Traditional mechanism synthesis follows two paths, both with unique challenges. Analytical methods rooted in kinematics, such as Burmester theory \cite{hartenberg1964kinematic}, produce closed-form solutions for simple four-bar linkages and a handful of target positions but break down for continuous curves, richer topologies, or cases with no solution \cite{primrose1964finite, ma1988performance}. Numerical optimization casts synthesis as error minimization inside a chosen topology \cite{ebrahimi2015efficient}, yet gradient methods cling to initial guesses and get trapped in local minima \cite{mariappan1996generalized, sancibrian2004gradient}, while population-based search needs thousands of evaluations \cite{lin2010ga, acharyya2009performance, cabrera2002optimal}. Both approaches usually explore only one topology, so better mechanisms remain hidden unless each alternative is optimized separately, multiplying cost.

Although finding a mechanism that produces a prescribed path is exceptionally difficult, simulating the path traced by a given mechanism is straightforward. This computational asymmetry between inverse and forward kinematics makes the problem well-suited to learning-based approaches. Because forward kinematics are inexpensive, thousands of candidate mechanisms can be evaluated in seconds on modern hardware, enabling large-scale dataset generation and screening \cite{heyrani2022links, nurizada2025dataset}.

Inspired by recent successes in domains like program synthesis \cite{wang2021codet5}, molecular design \cite{mazuz2023molecule}, and gear design \cite{etesam2025deep,cheong2025simft,ataei2025transformer}, we reframe kinematic synthesis as a conditional sequence-to-sequence learning problem. We introduce a domain-specific language (DSL) that serializes any mechanism into a structured string of tokens. This approach transforms the intractable design problem into a more manageable task: translating a target curve into its corresponding mechanism \emph{sentence}.

We present MechaFormer, a Transformer-based architecture specifically trained to perform this translation. Given a target B-spline curve, MechaFormer autoregressively generates a complete mechanism definition. Its core advantage lies in its unified approach: the model learns to output topology and geometry simultaneously, allowing it to discover the most appropriate type of mechanism for a given curve and fine-tune its parameters in a single, coherent process.

A major strength of our generative approach is its capacity for interactive and exploratory design, overcoming the limitations of single-point predictions. We introduce and validate a suite of novel sampling strategies that enable a more holistic exploration of the vast and complex design space. These techniques empower designers to rapidly generate a diverse portfolio of candidate solutions, systematically probing variations in mechanism topology, geometry, and even contextual placement within a given environment. By providing a broader view of the possibilities, our method allows for the discovery of superior and non-obvious solutions.

This paper makes the following contributions:
\begin{enumerate}
    \item \textbf{A unified generative framework} that treats mechanism synthesis as a sequence learning problem. This approach seamlessly integrates topology selection and parameter optimization into a single, end-to-end model, allowing it to discover optimal mechanisms from a vast and complex design space without manual partitioning.
    \item \textbf{State-of-the-art performance} in path synthesis with sampling supporting different topologies. Through comprehensive experiments, we show that MechaFormer surpasses existing learning-based baselines in path-matching accuracy, while also generating a diverse set of valid and novel mechanism designs.
    \item \textbf{A design framework with targeted sampling strategies.} We introduce and validate sampling strategies, including Best@$k$, rotational sampling, and topology sampling that substantially improve solution quality and diversity, enabling rapid exploration of candidate mechanisms.
    \item \textbf{A hybrid optimization workflow.} We show that model outputs provide superior initializations for local optimizers, yielding elite solutions far more efficiently and robustly than direct optimization from random starts.
\end{enumerate}

\section{Related Work}
Recent contributions in mechanical synthesis of linkage mechanisms have looked to machine learning techniques for guidance on designing complex physical systems \cite{sonntag2024machine, han2025review}.

Nurizada et al. (\citeyear{nurizada2025path}) introduced a conditional $\beta$-VAE model for generating diverse planar four-bar mechanisms from coupler curves. Their work is supported by a large curated dataset and a comprehensive evaluation framework based on reconstruction accuracy, novelty, and diversity. While the model is capable of producing multiple candidate mechanisms per input, its reconstruction performance shows only about 45\% of generated mechanisms achieve a DTW score below 2, which indicates satisfactory curve approximation. This limits its effectiveness for high-precision synthesis tasks and suggests room for improvement in learning the mapping between input curves and mechanism parameters.

Nobari et al. (\citeyear{heyrani2022links}) introduced the LINKS dataset, a large-scale collection of 100 million planar mechanisms ranging from 6 to 15 links, each paired with a motion trajectory. In follow-up work, Nobari et al. (\citeyear{nobari2024link}) developed a contrastive learning approach for retrieving mechanisms from the dataset, followed by local refinement using quasi-Newton methods. While effective in minimizing trajectory error, many of the retrieved mechanisms are topologically complex and structurally unrealistic, often involving oversized configurations or internal collisions that would be impractical or impossible to manufacture. 

Optimization and reinforcement learning-based methods have also been explored for mechanism synthesis. Pan et al. (\citeyear{pan2023joint}) investigated a range of optimization approaches, including mixed-integer conic programming (MICP), mixed-integer nonlinear programming (MINLP), and simulated annealing (SA), to design mechanisms with up to 14 rigid links. Their hybrid MICP-MINLP approach achieved higher-quality results than SA but incurred high computational costs, often requiring over an hour to solve a single instance. Similarly, Fogelson et al. (\citeyear{fogelson2023gcp}) proposed a reinforcement learning framework that refines mechanism parameters through reward-based feedback while enforcing geometric constraints. Although their method demonstrates better performance than evolutionary baselines, it remains expensive and requires extensive reward engineering to balance feasibility and task performance. These approaches illustrate the promise and limitations of optimization-heavy pipelines, which struggle to scale efficiently.

In contrast to prior work, our approach treats mechanism synthesis as a sequence modeling task, using a transformer-based architecture to map continuous B-spline representations of input curves to discrete DSL tokens representing mechanism topology and joint parameters. This formulation enables generalization across a diverse set of mechanism families and allows for probabilistic sampling of multiple topologies and configurations per input. Furthermore, we demonstrate how this framework can be used to discover better-performing topologies through input rotation and topology sampling, a form of post-hoc analysis that has not been explored in existing generative models. 

\section{Problem Formulation}

The core task of kinematic path synthesis is to solve an inverse problem. Given a desired target trajectory, the goal is to discover a corresponding planar linkage mechanism that can generate it. We formalize the key components below.
We represent a mechanism as $\mathcal{M}=(\tau,\mathbf{J},c)$, where $\tau\in\mathcal{T}$ is a topology chosen from a finite set of mechanism types (e.g., four-bar, Watt six-bar) that fixes the number of links, the joint types, and their connectivity (revolute or prismatic). The continuous parameters are $\mathbf{J}={\mathbf{j}_1,\ldots,\mathbf{j}_n}\subset\mathbb{R}^2$, the initial joint coordinates for $\tau$. The coupler index $c$ selects the joint whose path must match the target.

Forward kinematics is given by a deterministic simulator $\Phi$ that maps $\mathcal{M}$ and an actuation variable (e.g., input crank angle $\theta$) to the coupler position and thus to the curve
\begin{equation}
\mathcal{C}=\Phi(\mathcal{M})={\Phi(\mathcal{M},\theta)\mid \theta\in[0,2\pi)}\subset\mathbb{R}^2.
\end{equation}
This forward process, $\Phi$, is computationally inexpensive and straightforward to evaluate.

The inverse problem (finding an optimal mechanism $\mathcal{M}^*$ that generates a trajectory closely matching a desired target curve $\mathcal{C}^*$) can be framed as an optimization problem:
\begin{equation}
    \mathcal{M}^* = \underset{\mathcal{M} = (\tau, \mathbf{J}, c)}{\arg\min} \  d(\Phi(\mathcal{M}), \mathcal{C}^*)
    \label{eq:optimization_problem}
\end{equation}
where $d(\cdot, \cdot)$ is a distance metric between two curves, such as Dynamic Time Warping (DTW) \cite{JMLR:v21:20-091}. This formulation is notoriously difficult to optimize because it is both mixed-integer and non-convex: the search couples a discrete topology $\tau \in \mathcal{T}$ with continuous joint coordinates $\mathbf{J} \in \mathbb{R}^{2n}$, and the objective $d(\Phi(\mathcal{M}), \mathcal{C}^*)$ has many local minima in $\mathbf{J}$, making gradient-based and many gradient-free methods unreliable and highly sensitive to initialization.

To address these challenges, we reframe kinematic synthesis as a probabilistic modeling task rather than optimization. We learn the conditional distribution $p(\mathcal{M} | \mathcal{C}^*)$ from a dataset of $(\mathcal{M}, \mathcal{C})$ pairs generated by the forward simulator $\Phi$, enabling direct sampling of high-quality mechanisms for any target curve.

\section{Methodology}

\subsection{Canonical Frame Normalization for Invariance and Exploratory Sampling}
A fundamental challenge in learning from geometric data is handling nuisance variables, in our case a mechanism's function is invariant to similarity transformations. Raw joint coordinates force models to waste capacity learning this invariance. We eliminate this redundancy by applying canonical normalization to every mechanism, anchoring on its two ground joints $\mathbf{j}_{g_1}$ and $\mathbf{j}_{g_2}$. An affine transformation maps all joints such that $\mathbf{j}'_{g_1} = (0,0)$ and $\mathbf{j}'_{g_2} = (1,0)$.

\begin{figure*}[ht]
    \centering
    \includegraphics[width=\linewidth]{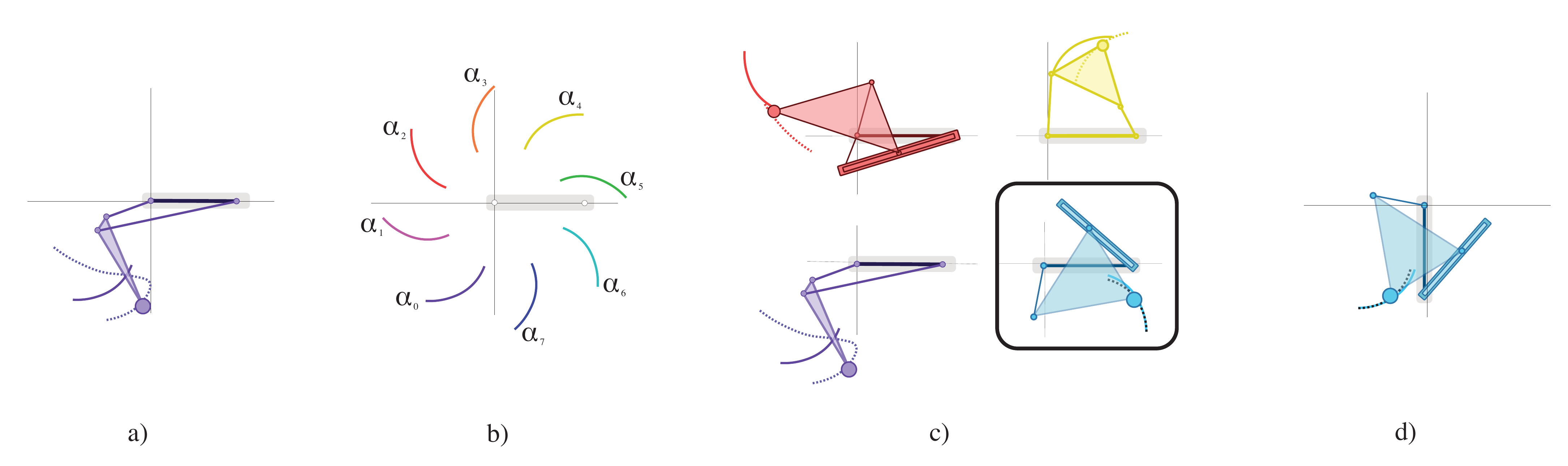}
    \caption{
        a) User-defined input curve shown by the solid line, and model-generated mechanism and coupler trajectory shown by the dashed line. 
        b) Original curve is rotated eight times by $\Delta\alpha = 45^\circ$. 
        c) Four example mechanisms at different rotation angles. The mechanism at $\alpha_6$ is selected. 
        d) The mechanism base is rotated to retain the relative position of the user-defined input curve relative to the origin. A simple target curve is used to clearly illustrate the process.
    }
    \label{fig:rotation-full}
\end{figure*}

This normalization provides two key benefits. First, it collapses infinite placements to one canonical representation, letting the model focus on mapping relative geometry to curve shape. Second, by fixing the mechanism's frame, we enable design exploration through input curve transformation, the foundation of our rotational sampling strategy discussed later (Figure \ref{fig:rotation-full}). To find optimal mechanism orientation for a given curve, we rotate the input by angles $\{\alpha_1, \dots, \alpha_m\}$, generate mechanisms for each rotation, then apply inverse rotations to the generated mechanisms. This efficiently searches for optimal base orientation in real-world coordinates, made tractable by our representation.

\subsection{Abstracting Topology as a Single Conditional Choice}
We represent mechanism topology as a single categorical variable rather than having the model assemble a graph of links and joints. Each of the 24 valid, kinematically distinct topologies in our library maps to a unique token (e.g., \texttt{RRRR}, \texttt{Watt2T1A1}, \texttt{Steph1T1}), and the model’s first task is to predict one token.

This abstraction guarantees kinematic feasibility and focuses learning where it matters. Constructing valid chains directly is combinatorially hard and constrained by rules such as Grubler’s criterion \cite{gogu2005chebychev} and connectivity, so encoding topology as a token ensures every design starts from a valid, well-defined type. With topology fixed, the model maps curve features to the most suitable known mechanism, then infers dimensions. The detailed connectivity graph for the chosen type is stored externally and guides generation of joint coordinates.

This choice does preclude novel topologies, which is an acceptable trade-off. Most practical mechanisms use a small set of foundational four-bar and six-bar linkages that our 24 types cover well. The real engineering challenge is creative dimensioning within these proven families, and that is exactly what the model optimizes.

\subsection{Input Representation: B-Spline Control Points}
To provide the model with a compact, fixed-size representation of the target curve $\mathcal{C}^*$, we parameterize it using a cubic B-spline. Any given target curve, initially represented as a set of ordered points, is fitted to a B-spline with a fixed number of $K=64$ control points. The resulting set of control points, $\mathbf{P} = \{\mathbf{p}_1, \dots, \mathbf{p}_K\}$ with $\mathbf{p}_k \in \mathbb{R}^2$, captures the essential shape of the curve. This sequence of control points $\mathbf{X} \in \mathbb{R}^{K \times 2}$ serves as the input to our model.

\subsection{Output Representation: A Domain-Specific Language for Mechanisms}

To make the mechanism $\mathcal{M}$ amenable to sequence generation, we developed a DSL that serializes a mechanism's topology and continuous joint coordinates into a discrete sequence of tokens, $\mathbf{y} = (y_1, y_2, \dots, y_L)$.

\paragraph{Vocabulary} The vocabulary $\mathcal{V}$ of our DSL consists of three types of tokens:

\begin{itemize}
    \item \textbf{Structural Tokens:} Special tokens like `SOS' (Start of Sequence), `EOS' (End of Sequence), and delimiters like `MECH\_TYPE', `POINTS', `P\_i', `X:', `Y:'.
    \item \textbf{Topology Tokens:} A finite set of tokens representing each of the 24 mechanism topologies (including 4-bar and 6-bar linkages, with a combination of revolute or prismatic joints) in our design space, e.g., `RRRR', `RRRP'.
    \item \textbf{Coordinate Tokens:} A set of integer tokens representing quantized coordinate values. We discretize the continuous coordinate space $[-\kappa, \kappa]$ into $B=200$ uniform bins (see appendix for the ablation that justified this choice). A continuous coordinate value $v$ is mapped to a bin index $b(v)$ via:
    \begin{equation}
        b(v) = \left\lfloor \frac{(v + \kappa) \cdot (B-1)}{2\kappa} \right\rfloor
    \end{equation}
    This transforms the regression of continuous joint positions into a multi-class classification problem.
\end{itemize}

\paragraph{Sequence Structure} A full mechanism sequence $\mathbf{y}$ follows a strict, human-readable structure that first declares the topology and then lists the normalized coordinates of its free joints. A ground link is assumed to be normalized with its joints at $(0,0)$ and $(1,0)$, so these are not included in the sequence. Table \ref{tab:dsl_breakdown} provides a detailed example.

\begin{table}[h!]
\centering
\small
\renewcommand{\arraystretch}{1.2}
\begin{tabular}{p{0.25\linewidth} p{0.65\linewidth}}
\toprule
\textbf{Token Sub-Sequence} & \textbf{Description} \\
\midrule
\texttt{<SOS>} & Start of sequence token. \\
\texttt{MECH\_TYPE} \texttt{RRRR} & Declares the mechanism's topology. Here, a four-bar linkage with four revolute joints. \\
\texttt{POINTS} & Structural token indicating the start of the joint coordinate block. \\
\texttt{P\_1 X: BIN\_125 Y: BIN\_180} & Specifies the coordinates for the first free joint. The continuous coordinates $(x_1, y_1)$ have been quantized to integer bin indices (125, 180). \\
\texttt{P\_2 X: BIN\_45 Y: BIN\_78} & Specifies the coordinates for the second free joint, quantized to bins (45, 78). \\
... & (Additional joints) \\
\texttt{<EOS>} & End of sequence token. \\
\bottomrule
\end{tabular}
\caption{Breakdown of the Domain-Specific Language (DSL) used to represent a mechanism as a token sequence.}
\label{tab:dsl_breakdown}
\end{table}

\subsection{Sequence-to-Sequence Model Architecture}
We map an input sequence of control points $\mathbf{X}$ to a sequence of mechanism tokens $\mathbf{y}$ using a Transformer encoder-decoder \cite{vaswani2017attention}. Training maximizes
\begin{equation}
\log p(\mathbf{y}\mid \mathbf{X})=\sum_{t=1}^{L}\log p(y_t\mid \mathbf{y}_{<t},\mathbf{X}).    
\end{equation}

The encoder self-attends over the B-spline control points to obtain contextual features. The decoder autoregressively predicts tokens using masked self-attention over $\mathbf{y}{<t}$ and cross-attention to the encoder output.

\subsection{Hybrid Method: Generative Seeding for Local Optimization}

MechaFormer generates high-quality mechanisms but may not reach perfect local optima due to its probabilistic nature. We bridge this gap by combining our generative model's global search with local optimization precision, using MechaFormer's output as an intelligent initial seed.

For target curve $\mathcal{C}^*$, we first sample an initial mechanism $\mathcal{M}_{gen} = (\tau_{gen}, \mathbf{J}_{gen}, c_{gen})$ from $p(\mathcal{M} | \mathcal{C}^*)$. This handles the most difficult aspects: selecting a promising topology $\tau_{gen}$ and providing near-optimal initial geometry $\mathbf{J}_{gen}$.

We then fix the topology $\tau_{gen}$ and refine the joint coordinates through local optimization. Within trust region $\mathcal{B}$ around $\mathbf{J}_{gen}$, we minimize path-following error:
\begin{equation}
    \mathbf{J}^* = \underset{\mathbf{J} \in \mathcal{B}(\mathbf{J}_{gen})}{\arg\min} \  d(\Phi((\tau_{gen}, \mathbf{J}, c_{gen})), \mathcal{C}^*)
\end{equation}

We solve this using L-BFGS-B, a quasi-Newton method with box constraints. By starting from MechaFormer's high-likelihood initialization rather than random points, this hybrid approach avoids poor local minima and consistently achieves superior solutions.

\subsection{Dataset and Training}
We train MechaFormer on a subset of the dataset introduced by \citet{nurizada2025dataset}, which contains 3 million single-DOF planar linkage mechanisms with their corresponding coupler curves. We filter the dataset to include only mechanism types with at least 20,000 instances, resulting in 846,480 training samples (with 83,499 held out for validation) across 24 distinct topologies. Our model architecture contains approximately 19 million parameters with a 256-dimensional hidden size, 8 attention heads, and 6 layers each for encoder and decoder. The model is trained using the AdamW optimizer with a cosine learning rate schedule and warm-up, employing cross-entropy loss with label smoothing for 30 epochs on distributed GPUs (DGX A100). Implementation details, including the training code, processed data, and model weights will be provided with the final version.

\section{Experiments}

\begin{table*}[t]
\centering
\small
\begin{tabular}{
    >{\centering\arraybackslash}p{0.9cm}  
    p{2.3cm} 
    >{\centering\arraybackslash}p{1.5cm}
    >{\centering\arraybackslash}p{1.7cm}
    >{\centering\arraybackslash}p{1.7cm}
    >{\centering\arraybackslash}p{1.9cm}
    >{\centering\arraybackslash}p{1.5cm}
    >{\centering\arraybackslash}p{1.9cm}
}
\hline\hline \\[-1.75ex]
& \textbf{} 
& $\textnormal{$\eta$}_{\small \mathrm{DTW}} \downarrow$ 
& $\textnormal{$\mu$}_{\small \mathrm{DTW}} \downarrow$ 
& DTW $< 3.0$ 
& DTW $< 2.0$ 
& $\textnormal{$\mu$}_{\small \mathrm{CD}} \downarrow$ 
& Success Rate $\uparrow$ \\
\specialrule{0.01em}{.1em}{.5em}

\multirow{9}{*}{\parbox[c][11.5ex][c]{0.9cm}{\centering\rotatebox[origin=c]{90}{\textbf{MechaFormer}}}}
& \textbf{Best @ $k$} & & & \\
& \begin{tabular}[t]{c}$k = 1$\\$k = 2$\\$k = 4$\\$k = 8$\\$k = 16$\\$k = 32$\end{tabular} &
  
  \begin{tabular}[t]{c}3.090\\
  2.652\\
  2.154\\
  1.877\\
  1.735\\
  \textbf{1.605}
  \end{tabular} &

\begin{tabular}[t]{c}
6.831\raisebox{0.5ex}{\scriptsize$\pm$9.468} \\
5.357\raisebox{0.5ex}{\scriptsize$\pm$6.699} \\
4.291\raisebox{0.5ex}{\scriptsize$\pm$5.499} \\
3.630\raisebox{0.5ex}{\scriptsize$\pm$4.848} \\
3.073\raisebox{0.5ex}{\scriptsize$\pm$3.609} \\
2.739\raisebox{0.5ex}{\scriptsize$\pm$3.212} \\
\end{tabular}
  
  &

\begin{tabular}[t]{c} 48.7$\%$\\
 54.6$\%$\\
61.9$\%$\\
66.2$\%$\\
69.5$\%$\\
\textbf{72.4}$\%$
\end{tabular} &

\begin{tabular}[t]{c} 35.3$\%$\\
41.8$\%$\\
47.4$\%$\\
52.3$\%$\\
56.1$\%$\\
\textbf{60.1}$\%$
\end{tabular} &

  \begin{tabular}[t]{c}0.250\\
  0.205\\
  0.171\\
  0.152\\
  0.134 \\
  0.123 
  \end{tabular} &

\begin{tabular}[t]{c}
  99.2\%\\
  99.4\%\\
  \textbf{99.5}\%\\
  \textbf{99.5}\%\\
  99.3\%\\
  99.4\%
\end{tabular} \\[-2.0ex]

& \textbf{Best @ $\alpha$} 
& 1.757 
& \hspace{.65em}\textbf{2.675\raisebox{0.5ex}{\scriptsize$\pm$2.534} }
& 72.0\% 
& 55.6\% 
& 0.121 
& 99.0\%\\

& \textbf{Best @} Topology 
& 2.090 
& \hspace{.65em}2.857\raisebox{0.5ex}{\scriptsize$\pm$2.468} 
& 68.4\% 
& 47.6\% 
& \textbf{0.119}  
& 86.7\% \\
\midrule

\multirow{4}{*}{\parbox[c][6.2ex][c]{0.9cm}{\centering\rotatebox[origin=c]{90} {\shortstack{\textbf{MF + }\\\textbf{BFGS}}} } }
& \textbf{Hybrid @ $k$} & & & & \\
& \begin{tabular}[t]{c}$k = 1$\\$k = 16$\\$k = 32$\end{tabular} &
  \begin{tabular}[t]{c}1.591\\0.912\\\textbf{0.887}\end{tabular} &
    \begin{tabular}[t]{c}
      4.264\raisebox{0.5ex}{\scriptsize$\pm$6.133}\\
      1.946\raisebox{0.5ex}{\scriptsize$\pm$2.535}\\
\textbf{      1.796\raisebox{0.5ex}{\scriptsize$\pm$2.349}
}    \end{tabular} 
    &
  \begin{tabular}[t]{c}65.0$\%$\\79.6$\%$\\\textbf{82.0}$\%$\end{tabular} &                 
  \begin{tabular}[t]{c}56.0$\%$\\71.7$\%$\\\textbf{73.9}$\%$\end{tabular} &

  \begin{tabular}[t]{c}0.139\\0.084\\\textbf{0.077}\end{tabular} &
  
  \begin{tabular}[t]{c}\textbf{97.6}\%\\94.4\%\\93.4\%\end{tabular} \\
\midrule

\\[-2.5ex]
\multirow{3.5}{*}{\parbox[c][7.2ex][c]{0.3cm}{\centering\rotatebox[origin=c]{90} {\textbf{Baseline}} } } \\[-1.2ex] 

& \textbf{KNN} & 3.799 & 5.356\raisebox{0.5ex}{\scriptsize$\pm$6.311} & 36.7\% & 20.7\% & 0.212 & 100.0\% \\

& \textbf{L-BFGS-B } & 14.258 & 18.430\raisebox{0.5ex}{\scriptsize$\pm$18.504} & 6.8\% & 3.9\% & 0.6603 & 95.0\% \\
\\

\hline\hline
\\[-2.0ex]
\multicolumn{8}{r}{\small \text{Prior work reports a best $\textnormal{$\mu$}_{\small \mathrm{CD}}$ of 0.135 from a study conducted on this dataset \cite{nurizada2025dataset}.}} \\
\\[-2.0ex]
\cline{5-8}

\end{tabular}
\caption{Performance comparison of sampling strategies (varying $k$, rotation angle $\alpha$, topology search) versus optimization baselines. DTW measures reconstruction accuracy: $<2$ (satisfactory), $2-3$ (moderate), $>3$ (poor). Both median and mean DTW reported due to high-variance outliers. Mean Chamfer Distance reported as supplementary accuracy metric. All methods used identical hyperparameters (see appendix). Success Rate indicates percentage of kinematically feasible mechanisms.}
\label{tab:comparison_metrics}
\end{table*}

We evaluate model performance on 1,000 validation curves drawn from the dataset introduced by Nurizada et al. (\citeyear{nurizada2025dataset}). All evaluations were conducted with a model temperature of 0.1 (see appendix for sampling study). We use DTW to assess reconstruction fidelity, with values below 2 considered satisfactory, scores between 2 and 3 deemed moderate, and values above 3 indicative of poor reconstruction. DTW computations are performed using the \textit{tslearn} library \cite{JMLR:v21:20-091}.
 
The normalization used for DTW analysis across curves of different scales is defined as:
\begin{equation}
    \mathcal{N}(\mathcal{X}) = \frac{\mathcal{X} - \mu_i}{\sigma_{{RMS}_i}}
\end{equation}
where both input and output curves are normalized by the input curve's mean ($\mu_i$) and Root Mean Squared variance ($\sigma_{{RMS}_i}$), enabling fair comparison between curves of different scales. We report both medians ($\eta_{DTW}$) and means ($\mu_{DTW}$) since means are skewed by outliers (see appendix), in addition to a percentage of samples that fell within satisfactory scores. Success Rate is also presented to quantify convergence on feasible mechanism configurations. Lastly, we present the mean bi-directional Chamfer Distance ($\mu_{{CD}}$) as a supplementary accuracy metric. We calculate this metric by computing the average of the minimum Euclidean distances from each point on one curve to the nearest point on the other curve, in both directions.

To provide a more comprehensive visual understanding of performance, we include representative examples of input curves, generated mechanisms, and their corresponding evaluation metrics in the appendix.

\subsection{Best @ k}
We evaluate the benefit of sampling multiple candidates by generating $k \in \{1, 2, 4, 8, 16, 32\}$ mechanisms per validation curve and selecting the best based on DTW. Table \ref{tab:comparison_metrics} shows monotonic improvement as $k$ increases: median DTW decreases from 3.09 ($k$=1) to 1.61 ($k$=32), while maintaining over 99\% Success Rate throughout. This demonstrates that sampling multiple candidates significantly improves solution quality without sacrificing robustness.

\subsection{Rotational Sampling}
Our canonical normalization enables efficient exploration of mechanism orientations by rotating the input curve rather than the mechanism itself. We generate mechanisms for the input curve rotated at 45° increments ($\alpha_i = i \cdot 45°$ for $i = 0, \ldots, 7$), then apply inverse rotations to recover real-world placements. Table \ref{tab:comparison_metrics} shows this strategy improves median DTW from 3.09 to 1.83, demonstrating that optimal base orientation significantly impacts accuracy. Figure \ref{fig:rotation-full} illustrates this eight-rotation process, where selecting the best orientation (e.g., $\alpha_6$ in the example) yields superior curve-following performance. While translation sampling could be similarly implemented by shifting the input curve, we leave this exploration for future work.

\begin{figure}[h!]
    \centering
    \includegraphics[width=0.9\linewidth]{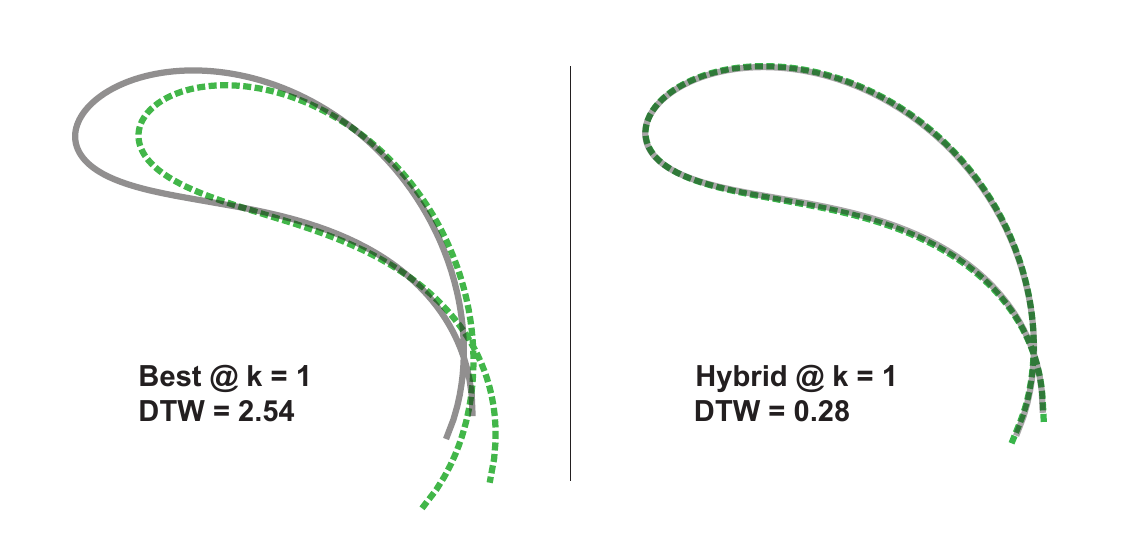}
    \caption{An example of curve alignment before and after implementing the optimization routine.}
    \label{fig:optimization_fig}
\end{figure}

\subsection{Topology Sampling}
Since topology tokens appear first in our DSL sequence, we can prefix-constrain generation to explore all 24 mechanism types for each input curve. Table \ref{tab:comparison_metrics} shows an 86.7\% Success Rate. Despite lower Success Rates, this sampling identifies the best-performing topology when design flexibility permits, as illustrated in Figure \ref{fig:fig1}.

\subsection{Hybrid Method}
We combine MechaFormer's Best @ $k$ sampling with local optimization: first generating $k$ mechanisms and selecting the best, then refining its joint coordinates using L-BFGS-B within trust regions. Table \ref{tab:comparison_metrics} shows this hybrid approach with $k=32$ achieves the lowest median DTW of 0.887, demonstrating that initializing from the best of multiple model samples dramatically outperforms random initialization. Figure~\ref{fig:optimization_fig} illustrates the improvement in DTW achieved through the use of hybrid optimization techniques.

\subsection{Baselines}
\subsubsection{Comparison to Prior Work}
We compare our approach against two recent works in mechanism synthesis. First, \citet{nurizada2025dataset} achieved a best $\mu_{\text{CD}}$ of 0.135. Our model surpasses this with $\mu_{\text{CD}}$ of 0.119 using sampling and 0.077 using our hybrid method (Table \ref{tab:comparison_metrics}). While both studies use subsets of the same dataset, direct comparison has limitations as the specific sample selections may differ. Second, \citet{nurizada2025path} reported $\eta_{\text{DTW}}$ of 2.441 using a conditional $\beta$-VAE on 12 million 4-bar mechanisms. Though our dataset is smaller (846,480 samples) and more diverse (24 topologies including 6-bar), the consistent DTW computation enables meaningful comparison. MechaFormer achieves $\eta_{\text{DTW}}$ of 1.605 (Best@32) and 0.887 (Hybrid@32), indicating that its architecture offsets reduced data scale while managing increased mechanism complexity.

\subsubsection{KNN}
To verify MechaFormer learns meaningful representations beyond memorization, we use the trained model's encoder to embed validation curves, retrieve their nearest neighbors from the training set based on cosine similarity in the embedding space, and evaluate the retrieved mechanisms. Table \ref{tab:comparison_metrics} shows MechaFormer significantly outperforms this baseline (median DTW: 3.090 vs 3.799), confirming the model generates novel solutions rather than retrieving stored examples from its learned representation space.
 
\subsubsection{L-BFGS-B}
We evaluate direct optimization by using \emph{just} the topology from MechaFormer's Best@k=32 output, applying the same ground normalization constraints ($\mathbf{j}'_{g_1} = (0,0)$, $\mathbf{j}'_{g_2} = (1,0)$), but randomly initializing all other joint coordinates. Using identical optimization parameters as the Hybrid study, Table \ref{tab:comparison_metrics} shows poor performance: median DTW of 14.258 and a 95\% convergence Success Rate, demonstrating the difficulty of optimization from random initialization in this non-convex design space.

\begin{table*}
\centering
\small
\begin{tabular}{lccccccccc}
\hline\hline \\[-1.75ex]
& \multicolumn{5}{c}{DTW (median $\pm$ std)} & \multicolumn{2}{c}{Success} & \multicolumn{2}{c}{Time} \\
\cmidrule(lr){2-6} \cmidrule(lr){7-8} \cmidrule(lr){9-10}
Approach & Initial & 25\% & 50\%  & 75\% & Final & DTW $<$ 3.0 & \texttt{maxfun} & \texttt{nfev} & (s) \\
\midrule
L-BFGS-B  
& 18.99\textsuperscript{\scriptsize ±11.26} 
& 11.87\textsuperscript{\scriptsize ±9.26} 
& 11.87\textsuperscript{\scriptsize ±9.26} 
& 11.86\textsuperscript{\scriptsize ±9.53} 
& 11.86\textsuperscript{\scriptsize ±9.53} 
& 3/10 & 0/10 
& 511\textsuperscript{\scriptsize ±316} 
& 2.22\textsuperscript{\scriptsize ±6.26} \\

Hybrid @ $k$ = 1 
& 4.36\textsuperscript{\scriptsize ±5.16} 
& 1.45\textsuperscript{\scriptsize ±2.26} 
& 1.43\textsuperscript{\scriptsize ±1.68} 
& 1.41\textsuperscript{\scriptsize ±1.61} 
& 1.15\textsuperscript{\scriptsize ±1.55} 
& 8/10 & 3/10 
& 641\textsuperscript{\scriptsize ±791} 
& 3.88\textsuperscript{\scriptsize ±17.73} \\

Hybrid @ $k$ = 16 
& 1.04\textsuperscript{\scriptsize ±5.19} 
& 0.77\textsuperscript{\scriptsize ±1.07} 
& 0.73\textsuperscript{\scriptsize ±0.94} 
& 0.71\textsuperscript{\scriptsize ±0.93} 
& 0.71\textsuperscript{\scriptsize ±0.94} 
& 9/10 & 2/10 
& 602\textsuperscript{\scriptsize ±672} 
& 3.54\textsuperscript{\scriptsize ±9.08} \\

Hybrid @ $k$ = 32 
& 0.99\textsuperscript{\scriptsize ±5.19} 
& 0.57\textsuperscript{\scriptsize ±4.69} 
& 0.53\textsuperscript{\scriptsize ±4.69} 
& 0.51\textsuperscript{\scriptsize ±4.31} 
& 0.51\textsuperscript{\scriptsize ±4.33} 
& 9/10 & 1/10 
& 567\textsuperscript{\scriptsize ±443} 
& 2.85\textsuperscript{\scriptsize ±8.80} \\
\hline\hline
\multicolumn{10}{l}{} \\
\end{tabular}
\caption{Optimization performance for 10 samples with \texttt{maxfun} set at 2000. The \texttt{nfev} is the number of objective evaluations used, and the \texttt{maxfun} indicates the number of runs that reached the evaluation cap. MechaFormer generation time is negligible.}\label{tab:optimization_analysis}
\end{table*}

\begin{figure*}
    \centering
    \includegraphics[width=0.8\linewidth]{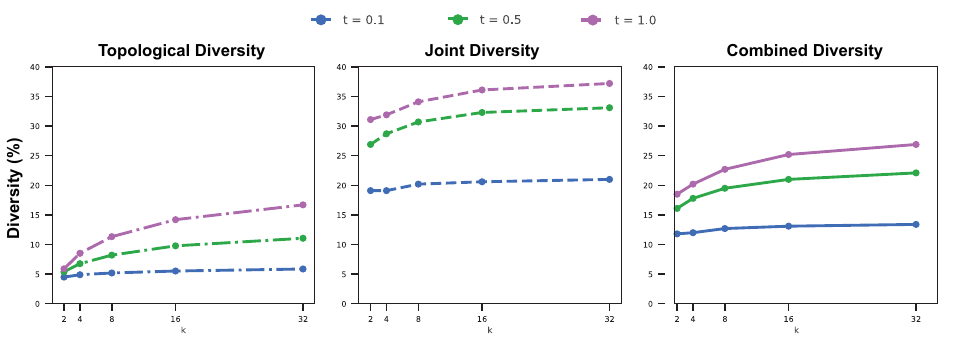}
    \caption{Topological, joint, and combined diversity trends across varying model temperatures and $k$-sampled evaluations. The Combined Diversity metric is the mean of the Topological and Joint Diversity values.}
    \label{fig:diversity_all}
\end{figure*}

\subsection{Diversity}
We quantify generative diversity to assess the model's ability to produce varied yet valid designs. For each input curve, we generate $k$ mechanisms and compute two diversity metrics. Topological diversity measures the fraction of unique mechanism types among $k$ samples, calculated as the number of unique topologies divided by the total 24 possible topologies. Joint diversity quantifies variation among mechanisms of the same topology by computing the average pairwise distance between their joint parameters, normalized by the square root of the number of free joints to account for varying mechanism complexities.

Figure \ref{fig:diversity_all} shows how temperature controls the exploration-exploitation tradeoff: $t=0.1$ yields consistent but limited variations (13.4\% combined diversity), while $t=1.0$ produces richer variations (26.9\%) at the potential cost of some invalid designs. This allows designers to tune between conservative refinement and creative exploration.

\subsection{Optimization Analysis}

To understand the value of model-guided initialization, we conduct a deeper analysis comparing L-BFGS-B  optimization from random starts versus our hybrid approach. We use SciPy's L-BFGS-B implementation with tolerance parameters \texttt{ftol}=$10^{-9}$ and \texttt{gtol}=$10^{-9}$, trust regions bounded by $\delta_i = \max(0.5, 0.5|x_i^0|)$, and extended optimization budget (\texttt{maxfun}=2000 vs. 100 initially, \texttt{maxiter}=50).

Table \ref{tab:optimization_analysis} shows results from 10 randomly selected validation samples. All methods achieve 90\% of their final improvement within 25\% of iterations, with median function evaluations ($\sim$600) well below the 2000 limit, indicating convergence to local optima rather than computational constraints. The critical finding is the 20× quality gap: L-BFGS-B  from random initialization achieves median DTW of 11.86 with only 30\% Success Rate ($DTW < 3.0$), while hybrid approaches reach 0.71 (k=16) to 0.51 (k=32) with 80-90\% success. This demonstrates that MechaFormer provides initializations in high-quality basins of attraction, and even with identical optimization parameters and extended budgets, random starts consistently converge to poor local minima due to the non-convex nature of the design space.

\section{Discussion and Conclusions}

MechaFormer demonstrates that reframing mechanism synthesis as conditional sequence generation yields substantial advantages over traditional approaches. Our DSL representation unifies topology selection and geometry optimization, eliminating exhaustive enumeration while enabling direct discovery of appropriate mechanisms from curve features.

Our sampling strategies unlock complementary benefits: Best@$k$ sampling reduces median DTW from 3.09 to 1.61, rotational sampling leverages canonical normalization to find optimal base orientations (DTW: 1.83), and temperature control enables designers to balance accuracy and diversity (13.4\% to 26.9\%). Most significantly, our hybrid approach achieves DTW of 0.887, over 20× better than direct optimization, by providing high-quality initializations that navigate the non-convex design space.

While limited to planar mechanisms with revolute/prismatic joints and a fixed topology library, MechaFormer establishes sequence learning as a powerful paradigm for mechanism design. Future work includes extending to spatial linkages, incorporating manufacturing constraints, and developing real-time CAD integration.

\bibliography{aaai2026}

\begin{thebibliography}{29}
\providecommand{\natexlab}[1]{#1}

\bibitem[{Acharyya and Mandal(2009)}]{acharyya2009performance}
Acharyya, S.; and Mandal, M. 2009.
\newblock Performance of EAs for four-bar linkage synthesis.
\newblock \emph{Mechanism and Machine Theory}, 44(9): 1784--1794.

\bibitem[{Angeles(2003)}]{angeles2003fundamentals}
Angeles, J. 2003.
\newblock \emph{Fundamentals of robotic mechanical systems: theory, methods, and algorithms}.
\newblock Springer.

\bibitem[{Ataei et~al.(2025)Ataei, Cheong, Jun, Matejka, Tessier, and Fitzmaurice}]{ataei2025transformer}
Ataei, M.; Cheong, H.; Jun, J.; Matejka, J.; Tessier, A.; and Fitzmaurice, G. 2025.
\newblock Transformer-Based Interfaces for Mechanical Assembly Design: A Gear Train Case Study.
\newblock \emph{arXiv preprint arXiv:2504.08633}.

\bibitem[{Bolanos et~al.(2023)Bolanos, Varela, Sargent, Stephen, Howell, and Magleby}]{bolanos2023selecting}
Bolanos, D.; Varela, K.; Sargent, B.; Stephen, M.~A.; Howell, L.~L.; and Magleby, S.~P. 2023.
\newblock Selecting and optimizing origami flasher pattern configurations for finite-thickness deployable space arrays.
\newblock \emph{Journal of Mechanical Design}, 145(2): 023301.

\bibitem[{Cabrera, Simon, and Prado(2002)}]{cabrera2002optimal}
Cabrera, J.; Simon, A.; and Prado, M. 2002.
\newblock Optimal synthesis of mechanisms with genetic algorithms.
\newblock \emph{Mechanism and machine theory}, 37(10): 1165--1177.

\bibitem[{Cheong et~al.(2025)Cheong, Ataei, Khasahmadi, and Jayaraman}]{cheong2025simft}
Cheong, H.; Ataei, M.; Khasahmadi, A.~H.; and Jayaraman, P.~K. 2025.
\newblock e-simft: Alignment of generative models with simulation feedback for pareto-front design exploration.
\newblock \emph{arXiv preprint arXiv:2502.02628}.

\bibitem[{Ebrahimi and Payvandy(2015)}]{ebrahimi2015efficient}
Ebrahimi, S.; and Payvandy, P. 2015.
\newblock Efficient constrained synthesis of path generating four-bar mechanisms based on the heuristic optimization algorithms.
\newblock \emph{Mechanism and Machine Theory}, 85: 189--204.

\bibitem[{Etesam et~al.(2025)Etesam, Cheong, Ataei, and Jayaraman}]{etesam2025deep}
Etesam, Y.; Cheong, H.; Ataei, M.; and Jayaraman, P.~K. 2025.
\newblock Deep generative model for mechanical system configuration design.
\newblock \emph{Proceedings of the AAAI Conference on Artificial Intelligence}, 39(16): 16496--16504.

\bibitem[{Fogelson, Tucker, and Cagan(2023)}]{fogelson2023gcp}
Fogelson, M.~B.; Tucker, C.; and Cagan, J. 2023.
\newblock GCP-HOLO: Generating high-order linkage graphs for path synthesis.
\newblock \emph{Journal of Mechanical Design}, 145(7): 073303.

\bibitem[{Gogu(2005)}]{gogu2005chebychev}
Gogu, G. 2005.
\newblock Chebychev--Gr{\"u}bler--Kutzbach's criterion for mobility calculation of multi-loop mechanisms revisited via theory of linear transformations.
\newblock \emph{European Journal of Mechanics-A/Solids}, 24(3): 427--441.

\bibitem[{Han et~al.(2025)Han, Zhao, Zhao, and Zi}]{han2025review}
Han, X.; Zhao, P.; Zhao, X.; and Zi, B. 2025.
\newblock Review on machine learning-based approaches for the kinematic analysis and synthesis of mechanisms.
\newblock \emph{Frontiers of Mechanical Engineering}, 20(2): 11.

\bibitem[{Hartenberg and Danavit(1964)}]{hartenberg1964kinematic}
Hartenberg, R.; and Danavit, J. 1964.
\newblock \emph{Kinematic synthesis of linkages}.
\newblock New York: McGraw-Hill.

\bibitem[{Heyrani~Nobari et~al.(2022)Heyrani~Nobari, Srivastava, Gutfreund, and Ahmed}]{heyrani2022links}
Heyrani~Nobari, A.; Srivastava, A.; Gutfreund, D.; and Ahmed, F. 2022.
\newblock Links: A dataset of a hundred million planar linkage mechanisms for data-driven kinematic design.
\newblock In \emph{International Design Engineering Technical Conferences and Computers and Information in Engineering Conference}, volume 86229, V03AT03A013. American Society of Mechanical Engineers.

\bibitem[{Lin(2010)}]{lin2010ga}
Lin, W.-Y. 2010.
\newblock A GA--DE hybrid evolutionary algorithm for path synthesis of four-bar linkage.
\newblock \emph{Mechanism and Machine Theory}, 45(8): 1096--1107.

\bibitem[{Ma and Angeles(1988)}]{ma1988performance}
Ma, O.; and Angeles, J. 1988.
\newblock Performance evaluation of path-generating planar, spherical and spatial four-bar linkages.
\newblock \emph{Mechanism and machine theory}, 23(4): 257--268.

\bibitem[{Mariappan and Krishnamurty(1996)}]{mariappan1996generalized}
Mariappan, J.; and Krishnamurty, S. 1996.
\newblock A generalized exact gradient method for mechanism synthesis.
\newblock \emph{Mechanism and Machine Theory}, 31(4): 413--421.

\bibitem[{Martins and Ning(2021)}]{martins2021engineering}
Martins, J.~R.; and Ning, A. 2021.
\newblock \emph{Engineering design optimization}.
\newblock Cambridge University Press.

\bibitem[{Mazuz et~al.(2023)Mazuz, Shtar, Shapira, and Rokach}]{mazuz2023molecule}
Mazuz, E.; Shtar, G.; Shapira, B.; and Rokach, L. 2023.
\newblock Molecule generation using transformers and policy gradient reinforcement learning.
\newblock \emph{Scientific Reports}, 13(1): 8799.

\bibitem[{Nobari et~al.(2024)Nobari, Srivastava, Gutfreund, Xu, and Ahmed}]{nobari2024link}
Nobari, A.~H.; Srivastava, A.; Gutfreund, D.; Xu, K.; and Ahmed, F. 2024.
\newblock Link: Learning joint representations of design and performance spaces through contrastive learning for mechanism synthesis.
\newblock \emph{arXiv preprint arXiv:2405.20592}.

\bibitem[{Norton and Han(2007)}]{norton2007design}
Norton, R.~L.; and Han, J. 2007.
\newblock \emph{Design of machinery}, volume~4.
\newblock McGraw-Hill Science/Engineering/Math.

\bibitem[{Nurizada et~al.(2025)Nurizada, Dhaipule, Lyu, and Purwar}]{nurizada2025dataset}
Nurizada, A.; Dhaipule, R.; Lyu, Z.; and Purwar, A. 2025.
\newblock A dataset of 3M single-DOF planar 4-, 6-, and 8-bar linkage mechanisms with open and closed coupler curves for machine learning-driven path synthesis.
\newblock \emph{Journal of Mechanical Design}, 147(4): 041702.

\bibitem[{Nurizada, Lyu, and Purwar(2025)}]{nurizada2025path}
Nurizada, A.; Lyu, Z.; and Purwar, A. 2025.
\newblock Path generative model based on conditional $\beta$-variational auto encoder for four-bar mechanism design.
\newblock \emph{Journal of Mechanisms and Robotics}, 17(6): 061004.

\bibitem[{Pan et~al.(2023)Pan, Liu, Gao, and Manocha}]{pan2023joint}
Pan, Z.; Liu, M.; Gao, X.; and Manocha, D. 2023.
\newblock Joint search of optimal topology and trajectory for planar linkages.
\newblock \emph{The International Journal of Robotics Research}, 42(4-5): 176--195.

\bibitem[{Primrose, Freudenstein, and Sandor(1964)}]{primrose1964finite}
Primrose, E.; Freudenstein, F.; and Sandor, G. 1964.
\newblock Finite Burmester theory in plane kinematics.
\newblock \emph{Journal of Applied Mechanics}, 31(4): 683--693.

\bibitem[{Sancibrian et~al.(2004)Sancibrian, Viadero, Garc{\i}a, and Fern{\'a}ndez}]{sancibrian2004gradient}
Sancibrian, R.; Viadero, F.; Garc{\i}a, P.; and Fern{\'a}ndez, A. 2004.
\newblock Gradient-based optimization of path synthesis problems in planar mechanisms.
\newblock \emph{Mechanism and machine theory}, 39(8): 839--856.

\bibitem[{Sonntag et~al.(2024)Sonntag, Br{\"u}njes, Luttmer, Corves, and Nagarajah}]{sonntag2024machine}
Sonntag, S.; Br{\"u}njes, V.; Luttmer, J.; Corves, B.; and Nagarajah, A. 2024.
\newblock Machine learning applications for the synthesis of planar mechanisms—a comprehensive methodical literature review.
\newblock In \emph{International Design Engineering Technical Conferences and Computers and Information in Engineering Conference}, volume 88414, V007T07A003. American Society of Mechanical Engineers.

\bibitem[{Tavenard et~al.(2020)Tavenard, Faouzi, Vandewiele, Divo, Androz, Holtz, Payne, Yurchak, Ru{\ss}wurm, Kolar, and Woods}]{JMLR:v21:20-091}
Tavenard, R.; Faouzi, J.; Vandewiele, G.; Divo, F.; Androz, G.; Holtz, C.; Payne, M.; Yurchak, R.; Ru{\ss}wurm, M.; Kolar, K.; and Woods, E. 2020.
\newblock Tslearn, A Machine Learning Toolkit for Time Series Data.
\newblock \emph{Journal of Machine Learning Research}, 21(118): 1--6.

\bibitem[{Vaswani et~al.(2017)Vaswani, Shazeer, Parmar, Uszkoreit, Jones, Gomez, Kaiser, and Polosukhin}]{vaswani2017attention}
Vaswani, A.; Shazeer, N.; Parmar, N.; Uszkoreit, J.; Jones, L.; Gomez, A.~N.; Kaiser, L.~u.; and Polosukhin, I. 2017.
\newblock Attention is All you Need.
\newblock In Guyon, I.; Luxburg, U.~V.; Bengio, S.; Wallach, H.; Fergus, R.; Vishwanathan, S.; and Garnett, R., eds., \emph{Advances in Neural Information Processing Systems}, volume~30. Curran Associates, Inc.

\bibitem[{Wang et~al.(2021)Wang, Wang, Joty, and Hoi}]{wang2021codet5}
Wang, Y.; Wang, W.; Joty, S.; and Hoi, S.~C. 2021.
\newblock Codet5: Identifier-aware unified pre-trained encoder-decoder models for code understanding and generation.
\newblock \emph{arXiv preprint arXiv:2109.00859}.

\end{thebibliography}

\clearpage


\appendix
\section*{Appendix: Supplementary Document}

\section{Implementation Details}

\paragraph{Data Preprocessing.}
The raw mechanism dataset\footnote{https://www.kaggle.com/datasets/purwarlab/four-six-and-eight-bar-mechanisms-with-curves} undergoes several preprocessing steps to create a suitable representation for sequence learning. Each mechanism in the dataset is first normalized to a canonical coordinate frame through a sequence of transformations that place the two ground (fixed) joints at standardized positions: the first at the origin $(0,0)$ and the second at $(1,0)$. This normalization eliminates the infinite variations due to translation, rotation, and scaling while preserving the essential kinematic relationships. For the target curves, we employ cubic B-spline fitting with a fixed number of 64 control points, providing a compact yet expressive representation that captures curve shapes while maintaining a consistent input dimension for the neural network. The continuous joint coordinates are discretized into 200 uniform bins spanning the range $[-10, 10]$, transforming the regression problem into a multi-class classification task that the Transformer can handle more effectively.

\paragraph{Domain-Specific Language.}
The DSL serializes each mechanism into a structured token sequence following a strict grammar. Each sequence begins with a start-of-sequence token, followed by the mechanism type declaration (e.g., \texttt{MECH\_TYPE: RRRR}), then a \texttt{POINTS:} marker, and finally the quantized coordinates of each free joint. Each joint is represented as \texttt{P\_i X: BIN\_j Y: BIN\_k}, where $i$ indexes the joint and $j,k$ are the bin indices for the x and y coordinates respectively. This structured format ensures that the model learns both the syntax and semantics of valid mechanism descriptions while maintaining interpretability.

\paragraph{Model Architecture and Training.}
MechaFormer employs a standard Transformer encoder-decoder architecture. The encoder takes as input a sequence of 64 B-spline control points (each point represented as 2D coordinates) and processes them through six self-attention layers to build a contextualized representation of the target curve. The decoder autoregressively generates the DSL token sequence, starting from a start-of-sequence token and producing one token at a time through six masked self-attention layers with cross-attention to the encoder output. Each token prediction is made from a vocabulary of 232 tokens (including special tokens, DSL structure tokens, mechanism type tokens, and 200 coordinate bins). The architecture incorporates several modern improvements including flash attention for computational efficiency, RMSNorm for training stability, gated linear units (GLU) with Swish activation in the feedforward layers, rotary positional embeddings for better length generalization, and QK normalization in attention layers. Table~\ref{tab:hyperparameters} summarizes all hyperparameters used in our experiments. Training is performed using distributed data parallel (DDP) across $8\times$ NVIDIA A100 GPUs. The training takes about one hour to complete. The complete training and inference code are made publicly available to facilitate reproducibility and future research.

\begin{table}[H]
\centering
\begin{tabular}{@{}ll@{}}
\toprule
\textbf{Hyperparameter} & \textbf{Value} \\
\midrule
\multicolumn{2}{l}{\textit{Model Architecture}} \\
Hidden dimension ($d_{model}$) & 256 \\
Attention heads & 8 \\
Encoder layers & 6 \\
Decoder layers & 6 \\
Total parameters & $\sim$19M \\
\midrule
\multicolumn{2}{l}{\textit{Training}} \\
Optimizer & Adam \\
Learning rate & $1 \times 10^{-4}$ \\
Weight decay & $1 \times 10^{-5}$ \\
Batch size (per GPU) & 256 \\
Total batch size & 2048 \\
Epochs & 30 \\
LR schedule & ReduceLROnPlateau \\
LR reduction factor & 0.5 \\
LR patience & 3 epochs \\
Gradient clipping & 1.0 \\
\midrule
\multicolumn{2}{l}{\textit{Loss Function}} \\
Loss type & Cross-entropy \\
Padding token ignored & \checkmark \\
\midrule
\multicolumn{2}{l}{\textit{Data Processing}} \\
B-spline control points & 64 \\
B-spline degree & 3 \\
Coordinate range & $[-10, 10]$ \\
Coordinate bins & 200 \\
Min. instances per type & 20,000 \\
Train/validation split & 90/10 \\
\midrule
\multicolumn{2}{l}{\textit{Architecture Features}} \\
Flash attention & \checkmark \\
RMSNorm & \checkmark \\
GLU feedforward & \checkmark \\
Rotary position embeddings & \checkmark \\
QK normalization & \checkmark \\
Swish activation & \checkmark \\
No bias in feedforward & \checkmark \\
\bottomrule
\end{tabular}
\caption{Hyperparameters used for training MechaFormer.}
\label{tab:hyperparameters}
\end{table}

\section{Coordinate Discretization Ablation}

To determine the optimal discretization granularity for joint coordinates, we conducted an ablation study varying the number of bins $B$ used to quantize the continuous coordinate space $[-10, 10]$. We evaluated three bin sizes: $B \in \{50, 200, 2000\}$, specifically chosen to represent coarse, medium, and fine discretization levels respectively. These values span two orders of magnitude to comprehensively assess how quantization resolution influences model accuracy.

The bin size directly influences model accuracy through two competing effects. With $B = 50$ (bin width = 0.4 units), the coarse quantization introduces substantial discretization error—each predicted coordinate can only take one of 50 possible values, limiting the precision with which joint positions can be specified. This manifests as poor reconstruction accuracy (DTW = 8.0628) since the model cannot place joints with sufficient precision to accurately trace the target curves.

At the opposite extreme, $B = 2000$ (bin width = 0.01 units) provides high spatial resolution but paradoxically yields the worst performance (DTW = 12.2427). This degradation occurs because fine-grained discretization creates a sparse, high-dimensional output space where each of the 2000 bins appears infrequently in the training data. The model struggles to learn robust patterns across this sparse categorical distribution, leading to poor generalization despite the theoretical capability for precise coordinate specification.

The optimal configuration at $B = 200$ (bin width = 0.1 units) balances these competing factors. It provides sufficient spatial resolution for accurate joint placement while maintaining a learnable output distribution where each bin appears frequently enough in the training data for the model to learn meaningful patterns. This finding guided our choice of $B = 200$ for all experiments reported in the main paper, achieving a median DTW of 3.0900 that represents nearly $3\times$ improvement over either extreme.

\begin{table}[h]
\centering
\begin{tabular}{ccc}
\hline
\textbf{Bin Size} & \textbf{DTW Median} \\
\hline
50   & 8.0628  \\
200  & \textbf{3.0900}  \\
2000 & 12.2427 \\
\hline
\end{tabular}
\caption{DTW median values for different bin sizes. We run these experiments with Best @ \textit{k} = 1 and 10 different samples.}
\label{tab:bin_ablation}
\end{table}

\section{Temperature Sampling}

The purpose of this study was to identify an appropriate sampling temperature \( T \) for generating mechanisms across experiments. Since temperature influences the stochasticity of the model, selecting a suitable value ensures both quality and diversity in generated outputs while avoiding performance degradation due to overly deterministic or overly random behavior. We evaluated the average best normalized DTW between the predicted and target trajectories across different temperature values \( T \in \{0.001, 0.1, 0.5, 1.0\} \) and sampling counts \( k \in \{1, 2, 4, 8\} \). As shown in Table~\ref{tab:temperature_study}, \( T = 0.1 \) achieved the best performance when \( k = 1 \). Furthermore, as \( k \) increased, \( T = 0.1 \) remained competitive, with DTW scores comparable to other temperatures. Based on this balance between quality and consistency, we selected \( T = 0.1 \) as the default temperature for all experiments.

\begin{table}[H]
\centering
\small
\begin{tabular}{lcccc}
\toprule
\textbf{Temperature} & \textbf{k = 1} & \textbf{k = 2} & \textbf{k = 4} & \textbf{k = 8} \\
\midrule
T = 0.001 & 6.8652 & 6.8652 & 6.8652 & 6.8652 \\
T = 0.1   & \textbf{6.8316} & 5.3570 & 4.2911 & 3.6301 \\
T = 0.5   & 7.5084 & \textbf{5.2166} & \textbf{3.8533} & \textbf{3.1371} \\
T = 1.0   & 9.6690 & 6.3901 & 4.5011 & 3.4234 \\
\bottomrule
\end{tabular}
\caption{Average best normalized DTW for different values of sampling temperature \( T \) and number of samples \( k \). Lower is better.}
\label{tab:temperature_study}
\end{table}

\section{Distributions}

\begin{figure}[h]
    \centering
    \includegraphics[width=0.80\linewidth]{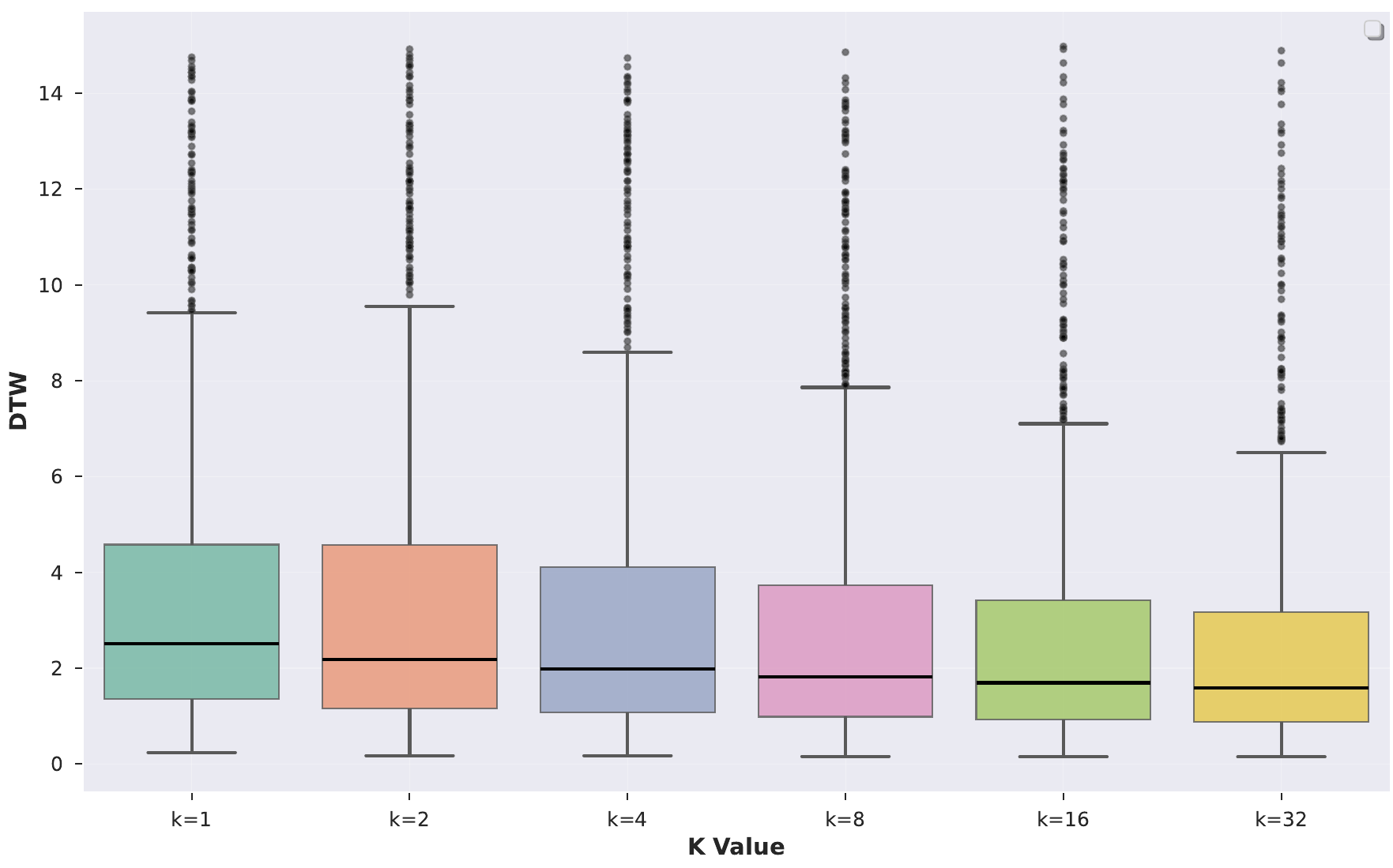}
    \caption{DTW distribution across different values of \( k \). Each box shows the interquartile range with the median marked. Extreme outliers are present and motivate comparisons across medians.}
    \label{fig:dtw_boxplot_linear}
\end{figure}

Figure~\ref{fig:dtw_boxplot_linear} shows the distribution of DTW scores for different sampling counts \( k \), plotted as a boxplot on a linear scale. Each box illustrates the interquartile range (IQR), with the central line indicating the median DTW value for each \( k \). We chose to highlight median values in our main text over means due to the presence of large outliers, which can significantly skew the average and misrepresent the typical performance. The boxplot makes this effect clear, particularly at lower \( k \) values where a few poorly performing samples inflate the upper range. The median provides a more robust summary statistic under these conditions, capturing the central tendency of the distribution more accurately.

\section{L-BFGS-B}

In this study, we use the best outcomes from the Best @ \textit{k} = 32 study, and search through the joint spaces for each of the free joints. This approach does not search through different topologies, as these are non-differentiable values. We initiate this problem using the following optimization statement:

\begin{align*}
\min_{\mathbf{x}} \quad & \text{DTW}(\mathcal{N}(\mathcal{X}_I), \mathcal{N}(\mathcal{X}_C(\mathbf{x}))) \\
\text{subject to} \quad & x_i \in [x_i^0 - \delta_i, x_i^0 + \delta_i], \quad i \in \{1,\ldots,n\} \\
& \mathbf{j}'_{g_1} = (0,0) \\
& \mathbf{j}'_{g_2} = (1,0) \\
\text{where} \quad & \delta_i = \max(0.5, 0.5|x_i^0|) , \\ 
& \mathcal{N}(\mathcal{X}) = \frac{\mathcal{X}_I - \mu_I}{\sigma_{{RMS}_I}}
\end{align*}

as defined by: 
$\mathbf{x}$ is the vector of mechanism coordinates to be optimized,
$\mathcal{X}_I$ is the input trajectory curve,
$\mathcal{X}_C(\mathbf{x})$ is the coupler trajectory for coordinates $\mathbf{x}$,
$\mathcal{N}(\cdot)$ is the normalization function,
$\mu$ is the mean of the input trajectory points,
$\sigma_{RMS_I}$ is the RMS variance of the input trajectory curve,
$x_i^0$ are the initial coordinate values,
$\delta_i$ are the bounds for each coordinate, and
$\mathbf{j}'_{g_1}$ and $\mathbf{j}'_{g_2}$ are fixed ground points at $(0,0)$ and $(1,0)$ respectively.
\\

\noindent This optimization routine is solved using the L-BFGS-B algorithm with the following parameters:
\begin{align*}
\texttt{maxiter} &= 50 \\
\texttt{maxfun} &= 100 \\
\texttt{ftol} &= 10^{-6} \\
\texttt{gtol} &= 10^{-6} \\
\texttt{eps} &= 10^{-3}
\end{align*}

\section{Examples}
\begin{figure*}[h]
    \centering
    \includegraphics[width=\linewidth]{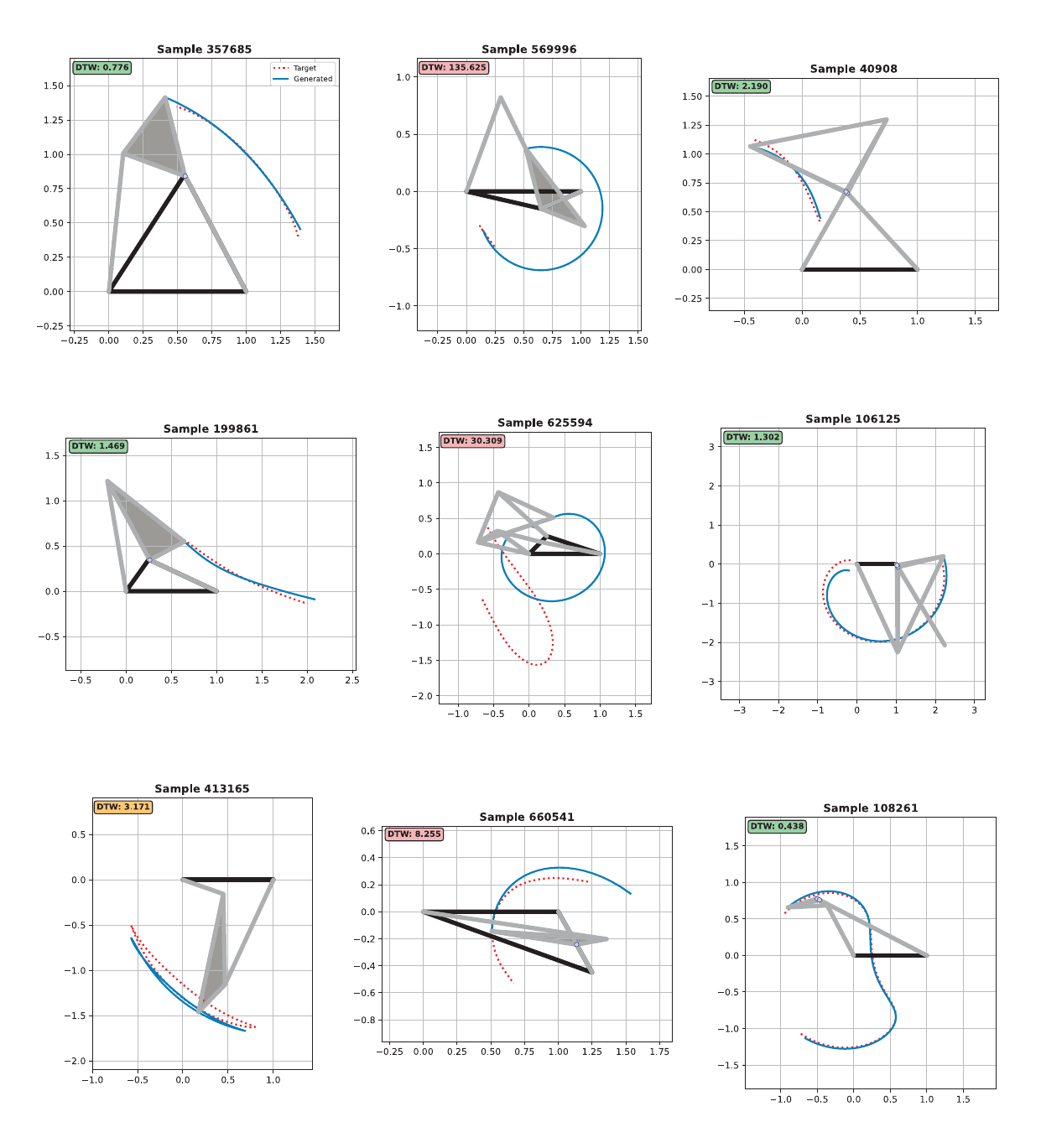}
    \caption{Nine sample mechanisms and their output trajectories in addition to the input curve. High, medium, and low accuracy performing outcomes are represented.}
    \label{fig:example-mechanisms}
\end{figure*}

\begin{figure*}[b]
     \centering
    \includegraphics[width=\linewidth]{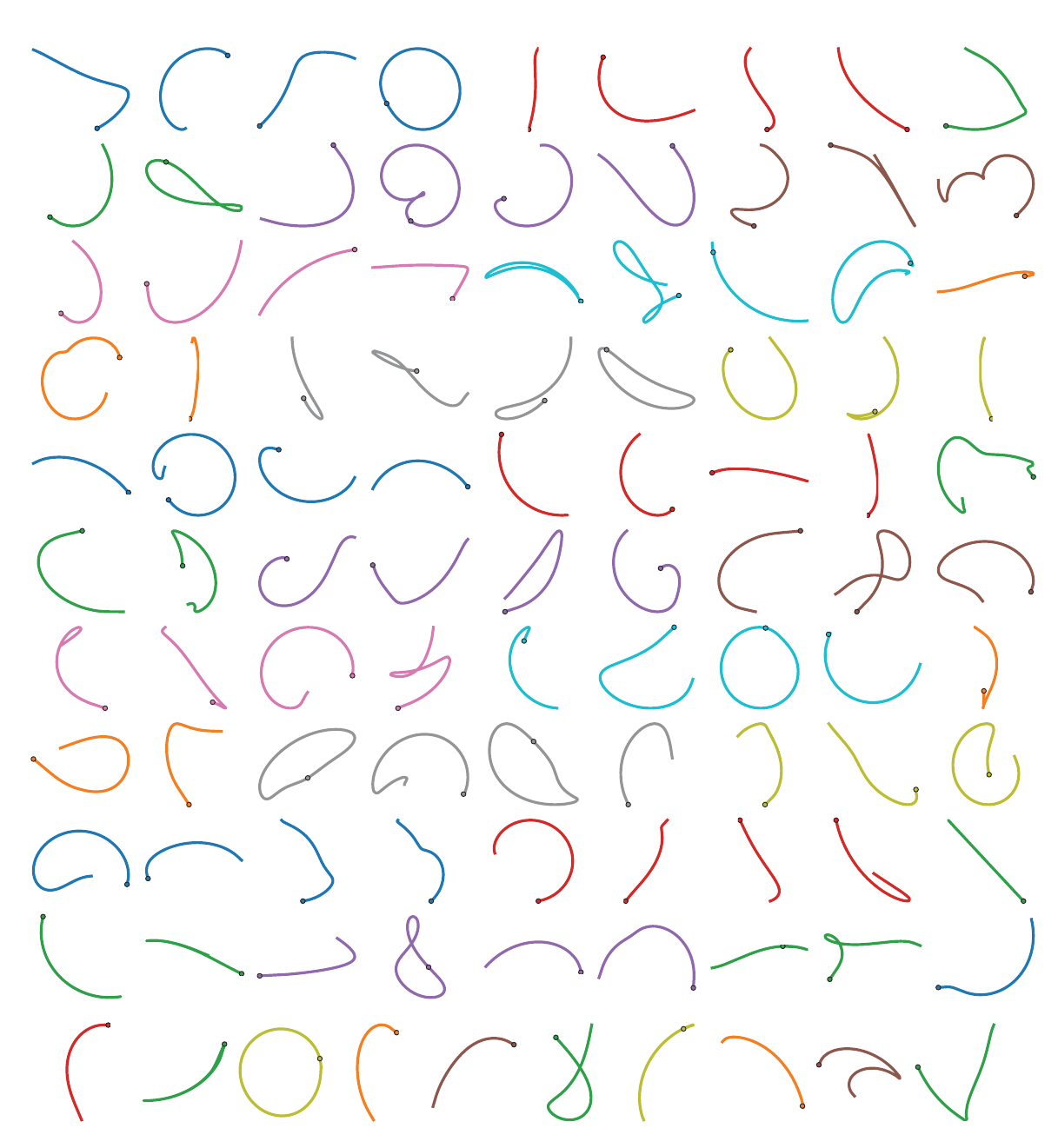}
    \caption{100 sample curves used in the dataset.}
    \label{fig:curves_100}   
\end{figure*}

Figure~\ref{fig:example-mechanisms} presents nine representative examples of generated mechanisms along with their corresponding target (input) and generated (output) trajectories. Each subplot displays a unique mechanism sample, visualizing the coupler path traced by the mechanism in relation to the desired curve. The DTW score is annotated in each plot, providing a quantitative measure of trajectory alignment. We deliberately selected a range of samples with varying performance levels to highlight the diversity in accuracy: from high-performing mechanisms with low DTW values (green) to low-performing ones with large trajectory mismatches (red). Figure \ref{fig:curves_100} shows 100 examples of different curves in the dataset used to train our model.

\end{document}